\documentclass[a4paper,conference]{IEEEtran}
\IEEEoverridecommandlockouts
\usepackage{cite}
\usepackage{graphicx}
\usepackage{amsmath,amssymb,mathtools,bm,etoolbox}
\usepackage{algorithmic}
\usepackage{color}
\usepackage{tikz}
\usetikzlibrary{matrix}
\usepackage{epstopdf}

\usepackage[labelformat=simple]{subcaption}

\usepackage{multirow}
\usepackage[ruled,vlined]{algorithm2e}
\usepackage{pgfplots}
\pgfplotsset{compat=1.14}
\pgfplotstableread{res/channels_kl.dat}{\channelskl}
\usepackage{url}
\usepackage{amsfonts}
\usepackage{soul}

\usepackage{textcomp}
\usepackage{xcolor}
\usepackage{hyperref}

\newcommand{\rk}[1]{#1}
\newcommand{\ua}[1]{{\color{gray}{}}}
\newcommand{\ff}[1]{{\color{black}{#1}}}

\def\BibTeX{{\rm B\kern-.05em{\sc i\kern-.025em b}\kern-.08em
    T\kern-.1667em\lower.7ex\hbox{E}\kern-.125emX}}
\begin{document}

\title{Epitomic Variational Graph Autoencoder}

% \author{
%     \IEEEauthorblockN{Author1\IEEEauthorrefmark{1}, Author2\IEEEauthorrefmark{2}, Author3\IEEEauthorrefmark{2}, Author4\IEEEauthorrefmark{1}}
%     \IEEEauthorblockA{\IEEEauthorrefmark{1}Institution1
%     \\\{1, 4\}@abc.com}
%     \IEEEauthorblockA{\IEEEauthorrefmark{2}Institution2
%     \\\{2, 3\}@def.com}
% }
\author{\IEEEauthorblockN{Rayyan Ahmad Khan\textsuperscript{\textsection}\IEEEauthorrefmark{1}
\IEEEauthorrefmark{2},
Muhammad Umer Anwaar\textsuperscript{\textsection}\IEEEauthorrefmark{1}\IEEEauthorrefmark{2}
,
Martin Kleinsteuber\IEEEauthorrefmark{1}\IEEEauthorrefmark{2}}
\IEEEauthorblockA{\IEEEauthorrefmark{1}
\textit{Technical University of Munich}\\
\{rayyan.khan, umer.anwaar, kleinsteuber\}@tum.de}
\IEEEauthorblockA{\IEEEauthorrefmark{2}
\textit{Mercateo AG}\\
Munich, Germany}
}
\maketitle
\begingroup\renewcommand\thefootnote{\textsection}
\footnotetext{Equal contribution}
\endgroup

\begin{abstract}
Variational autoencoder (VAE) is a widely used generative model for learning latent representations.
Burda et al. \cite{importance-weighted-autoencoders} in their seminal paper showed that learning capacity of VAE is limited by \textit{over-pruning}. 
It is a phenomenon where a significant number of latent variables fail to capture any information about the input data and the corresponding hidden units become inactive.
% Todo Umer: variables vs units vs factors. Which term to use? We have to be consistent if they mean the same thing. % hidden active, inactive units AND latent variables is finalized
This adversely affects learning diverse and interpretable latent representations. % and modeling power of VAE.
% Yeung et al. \cite{epitomic-vae} proposed epitomic VAE (EVAE) to solve this problem.
As variational graph autoencoder (VGAE) extends VAE for graph-structured data, it inherits the \textit{over-pruning} problem.
\ua{In this paper we propose a solution for this issue in VGAE. 
In addition, we identify the problem and a key reason of poor generative ability of VGAE.We adopt a model based approach and 
% motivated by the work of Yeung et al.\cite{epitomic-vae} on VAE. We 
propose epitomic VGAE (EVGAE), a generative variational framework for graph datasets which successfully mitigates the \textit{over-pruning} problem. }
\rk{In this paper, we adopt a model based approach and propose epitomic VGAE (EVGAE), a generative variational framework for graph datasets which successfully mitigates the \textit{over-pruning} problem and also boosts the generative ability of VGAE.}
We consider EVGAE to consist of multiple sparse VGAE models, called \textit{epitomes}, that are groups of latent variables sharing the latent space.
This approach aids in increasing active units as epitomes compete to learn better representation of the graph data. 
We verify our claims via experiments on three benchmark datasets. %, namely:  Cora, Citeseer and Pubmed.
Our experiments show that EVGAE has a better generative ability than VGAE. Moreover, EVGAE outperforms VGAE on link prediction task in citation networks.
\end{abstract}

% 8 pages inc references
\section{Introduction} 

Graphs are data structures that model data points via nodes and the relations between nodes via edges.
A large number of real world problems can be represented in terms of graphs. 
Some prominent examples are protein-protein interactions\cite{applications5}, social and traffic networks\cite{applications1,gcn} and knowledge graphs\cite{applications6}. % and economic activity flows \cite{applications8}.
% The following line can be skipped.
% Recently the field of deep learning has seen a noticeable growth in the interest in graph related problems because of increase in computational power, the power of graphs to model complex relations between objects and the ability of deep learning models to learn feature representations. 
Deep learning applications related to graphs include but are not limited to link prediction, node classification, clustering \cite{graph-survey1,graph-survey2} and recommender systems\cite{applications3,applications4,applications7}.
% in progress - quite non-linear way of proceeding - discuss with Rayyan
% Rephrase
% These applications make use of the techniques that learn node representations while making use of the information in both node features and the structure of the graph. 
% Make connection to previous thing and explain in a line the motivation of VGAE
% In the realm of unsupervised learning, variational graph autoencoder (VGAE)\cite{vgae} extends the variational autoencoder (VAE)\cite{vae} model to encode the input data into a low dimensional space.

\rk{
Kipf and Welling \cite{vgae} introduced variational graph autoencoder (VGAE) by extending the variational autoencoder (VAE) model \cite{vae}. Like VAE, VGAE tends to achieve the following two competing objectives:
\begin{enumerate}
	\item An approximation of input data should be possible.
	\item The latent representation of input data should follow standard gaussian distribution.
\end{enumerate}

There is, however, a well-known issue with VAE in general: The latent units, which fail to capture enough information about the input data, are harshly suppressed during training.
As a result the corresponding latent variables collapse to the prior distribution and end up simply generating standard gaussian noise.
Consequently, in practice, the number of latent units, referred to as \emph{active units}, actually contributing to reconstruction of the input data are quite low compared to the total available latent units. This phenomenon is referred to as \textit{over-pruning} (\cite{importance-weighted-autoencoders,over-pruning1,over-pruning2}). 
Several solutions have been proposed to tackle this problem for VAEs.
For instance, adding dropout can be a simple solution to achieve more active units. However, this solution adds redundancy rather than encoding more useful information with latent variables \cite{epitomic-vae}. \cite{over-pruning3} proposes division of the hidden units into subsets and forcing each subset to contribute to the KL divergence. \cite{over-pruning1} uses KL cost annealing to activate more hidden units. 
\cite{epitomic-vae} uses a model based approach where latent units are divided into subsets with only one subset penalized for a certain data point. 
These subsets also share some latent variables which helps in reducing the redundancy between different subsets. 

VGAE, being an extension of VAE for graph datasets, is also susceptible to the over-pruning problem. 
This greatly reduces the modeling power of pure VGAE and undermines its ability to learn diverse and meaningful latent representations As demonstrated in detail in Sec.~\ref{sec:overpruning}.
To suppress this issue, the authors of \cite{vgae} simply reduce the weight of the second objective by the number of nodes in training data. 
For instance, PubMed dataset\footnote{PubMed is a citation dataset\cite{citation-datasets}, widely used in deep learning for graph analysis. Details of the dataset are given in experiments Sec.~\ref{sec: datasets}} has $\sim$20k nodes, so the second objective is given 20,000 times less weight than the first objective.  Surprisingly, this factor is not mentioned in their paper, although it is present in their code \cite{vgae-to-gae-beta-factor}. 
Since the second objective is the one enforcing standard gaussian distribution for the latent variables, reducing its weight adversely affects the generative ability of VGAE and effectively reduces it to non-variational graph autoencoder. 
We discuss this further in Sec.~\ref{sec:gen_ability}. 

In this work, we refer to VGAE without any weighted objective as \emph{pure VGAE} to distinguish it from VGAE\cite{vgae}. In order to attain good generative ability and mitigate over-pruning, we adopt a model based approach called epitomic VGAE (EVGAE).
}
\ua{
Kipf and Welling \cite{vgae} introduced variational graph autoencoder (VGAE) by extending the variational autoencoder (VAE) model \cite{vae}.
% They employ graph convolutional network (GCN) \cite{gcn} as an encoder to learn the joint latent representation of nodes and edges. The decoder is simple inner product between latent variables.
Like the training objective of VAE, VGAE objective also consists of two competing terms. First term emphasizes learning better approximation of the input data. Second term ensures that the posterior distribution of the latent representation being learned is close to the prior distribution. We discuss this in detail in Sec.~II. 
In this work, we assume standard gaussian distribution as the prior distribution. 
% The latent representation of input data should follow standard gaussian distribution.

Unlike VAE, the training objective of VGAE 
proposed by Kipf and Welling 
\cite{vgae} contains an extra factor dividing the second term. This factor is the number of nodes in the graph. 
% Although 
Surprisingly, this factor is not mentioned in their paper, but it is present in their code \cite{vgae-to-gae-beta-factor}. 
This makes the second term quite small in comparison to the first term. For instance, PubMed dataset has $\sim$20k nodes, so the second term has 20,000 times less weightage than the first term.
% Assigning such an extremely low weight to the second term adversely 
This adversely affects the generative ability of VGAE. 
It effectively reduces it to non-variational graph autoencoder.
% The reason is that this term was responsible for ensuring that the posterior distribution being learned follows standard gaussian distribution. 
% Thus, when we generate new samples from standard gaussian distribution and pass it through the decoder, we get quite different output than the graph data used for training. 
We demonstrate this in Sec.~IV. 
% We also illustrate the relation of VGAE to $\beta$-VAE \cite{betavae1}.
% On Reduction of KL term  %TODO: see that correct section numbers are referred.
% TODO Rayyan: is "standard" compulsory in the modeling of VGAE?

Another problem that we address in this work is called \textit{over-pruning}.
It refers to the phenomenon when the majority of the latent units are not learning anything about the input data. 
The corresponding latent variables collapse to the prior distribution and are simply generating standard gaussian noise. 
It is a well-known issue in the training of VAE that 
% it is harsh in suppressing 
those latent units are harshly suppressed which fail to capture enough information about the input data. Consequently, the number of latent units, referred to as \emph{active units}, actually contributing to reconstruction of the input data are quite low  \cite{importance-weighted-autoencoders, over-pruning1, over-pruning2}.

Several solutions have been proposed to tackle this problem for VAEs.
For instance, adding dropout can be a simple solution to achieve more active units. However, this solution adds redundancy rather than encoding more useful information with latent variables \cite{epitomic-vae}. \cite{over-pruning3} proposes division of the hidden units into subsets and forcing each subset to contribute to the KL divergence. \cite{over-pruning1} uses KL cost annealing to activate more hidden units. 
\cite{epitomic-vae} uses a model based approach where latent units are divided into subsets with only one subset penalized for a certain data point. 
These subsets also share some latent variables which helps in reducing the redundancy between different subsets. 

VGAE, being an extension of VAE for graph datasets, is also susceptible to the over-pruning problem.
We refer to VGAE without any factor as \emph{pure VGAE} to distinguish it from VGAE\cite{vgae}.
We show in Sec.~III that pure VGAE indeed suffers from this problem.
This over-regularization greatly reduces the modeling power of pure VGAE and undermines its ability to learn diverse and meaningful latent representations.

In order to attain good generative ability and get rid of the over-pruning, we adopt a model based approach called epitomic VGAE (EVGAE).
% to address aforementioned two problems.
% Choose what to write here
% , motivated by \cite{epitomic-vae} . d
}
Our approach is motivated by a solution proposed for tackling over-pruning problem in VAE \cite{epitomic-vae}.
We consider our model to consist of multiple sparse VGAE models, called \textit{epitomes}, that share the latent space such that for every graph node only one epitome is forced to follow prior distribution. 
This results in a higher number of active units as epitomes compete to learn better representation of the graph data. 
% This enables other latent variables to be more free in encoding useful information for the node.
% We first give a brief overview of VAE and VGAE and look into over-pruning in more detail. 
% The section afterwards details epitopmic variational graph autoencoder (EVGAE). 
% In Sec.~VI, we validate the effectiveness of EVGAE via experiments on citation datasets \cite{citation-datasets}. Our experiments show that not only EVGAE achieves better generative ability than VGAE but also overcomes the over-pruning issue. 
Our main contributions are summarized below:
\begin{itemize}
    \item We identify that VGAE\cite{vgae} has poor generative ability due to the incorporation of weights in training objectives. % We relate VGAE to GAE and $\beta$-VAE.
    \item We show that pure VGAE (\ua{without any factor}\rk{without any weighted objectives}) suffers from the over-pruning problem.
    \item We propose a true variational model EVGAE that not only achieves better generative ability than VGAE but also mitigates the over-pruning issue.   
\end{itemize}

\section{Pure Variational Graph Autoencoder}
% VAE is one of the most widely used frameworks in deep learning. It assumes the observed data to have come from a joint distribution involving an unobserved(latent) variable. We aim to learn a representation of our data in a space of reduced dimensionality such that a reasonable reconstruction is possible and the assumptions on distributions are also met. 

Given an undirected and unweighted graph $\mathcal{G}$ consisting of $N$ nodes $\{\bm{x_1}, \bm{x_2}, \cdots, \bm{x_N}\}$ with each node having $F$ features. 
% We assume that the node embeddings follow some prior distribution. Specifically, 
We assume that the information in nodes and edges can be jointly encoded in a $D$ dimensional real vector space that we call latent space. We further assume that the respective latent variables $\{\bm{z_1}, \bm{z_2}, \cdots, \bm{z_N}\}$ follow standard gaussian distribution. 
These latent variables are stacked into a matrix $\bm{Z} \in \mathbb{R}^{N \times D}$. 
\ua{This matrix is then fed to a decoder network for reconstructing the input data.}
\rk{For reconstructing the input data, this matrix is then fed to the decoder network $p_{\theta}(\mathcal{G}| \bm{Z})$ parameterized by $\theta$.}
The assumption on latent representation allows the trained model to generate new data, similar to the training data, by sampling from the prior distribution.
Following VAE, the joint distribution can be written as
{\small
\begin{equation}
	p(\mathcal{G}, \bm{Z})  = p( \bm{Z}) p_{\theta}(\mathcal{G}| \bm{Z}),
\end{equation}
}%
where
{\small
\begin{align}
	p(\bm{Z})   & = \prod \limits_{i=0}^{N} p(\bm{z}_i) \label{eq:vgae-pz} \\                    
	p(\bm{z}_i) & = \mathcal{N}(\bm{0}, \mathrm{diag}(\bm{1})) \ \forall i .
	\label{eq:vgae-pz_i} 
\end{align}
}%
For an unweighted and undirected graph $\mathcal{G}$, we follow \cite{vgae} and restrict the decoder to reconstruct only edge information from the latent space.
The edge information can be represented by an adjacency matrix $\bm{A} \in \mathbb{R}^{N \times N}$ where \ua{$a_{ij}$} \rk{$\bm{A}[i, j]$} refers to the element in $i^{th}$ row and $j^{th}$ column. If an edge exists between node $i$ and $j$, we have \ua{$a_{ij} = 1$} \rk{$\bm{A}[i, j] = 1$}. 
Thus, the decoder is given by

\ua{
{\small
\begin{align}
	p_{\theta}(\bm{A}| \bm{Z}) = \prod \limits_{(i,j)=(1,1)}^{(N,N)}p_{\theta}(a_{ij} = 1| \bm{z_i}, \bm{z_j}), \label{eq:pA-given-Z}
\end{align}
}%
}
\rk{
	{\small
	\begin{align}
		p_{\theta}(\bm{A}| \bm{Z}) = \prod \limits_{(i,j)=(1,1)}^{(N,N)}p_{\theta}(\bm{A}[i, j] = 1| \bm{z_i}, \bm{z_j}), \label{eq:pA-given-Z}
	\end{align}
	}%	
}
with
\ua{
{\small
\begin{align}
	p_{\theta}(a_{ij} = 1| \bm{z_i}, \bm{z_j}) = \sigma(<\bm{z_i}, \bm{z_j}>),
\end{align}
}%
}
\rk{
{\small
\begin{align}
	p_{\theta}(\bm{A}[i, j] = 1| \bm{z_i}, \bm{z_j}) = \sigma(<\bm{z_i}, \bm{z_j}>),
\end{align}
}%

}
where $<.\ ,\  .>$ denotes dot product and $\sigma(.)$ is the logistic sigmoid function.
% Each component zi captures some latent source of variability in the data.

The training objective should be such that the model is able to generate new data
and recover graph information from the embeddings simultaneously. 
For this, \ua{we choose to maximize log probability of $\mathcal{G}$. i.e.}\rk{we aim to learn the free parameters of our model such that the log probability of $\mathcal{G}$ is maximized i.e. }
{\small
\begin{align}
	log\Big(p(\mathcal{G})\Big) & = log\Big(\int{ p( \bm{Z}) p_{\theta}(\mathcal{G}| \bm{Z}) \ d\bm{Z}}\Big) \nonumber                                                                                                    \\
	                            & = log\Big(\int{ \frac{q_{\phi}(\bm{Z} | \mathcal{G})}{q_{\phi}(\bm{Z} | \mathcal{G})} p( \bm{Z}) p_{\theta}(\mathcal{G}| \bm{Z}) \ d\bm{Z}}\Big) \nonumber                                            \\
	                            & = log\Big(\mathbb{E}_{\bm{Z} \sim q_{\phi}(\bm{Z} | \mathcal{G})} \Big\{\frac{p( \bm{Z}) p_{\theta}(\mathcal{G}| \bm{Z})}{q_{\phi}(\bm{Z} | \mathcal{G})}\Big\}\Big) , \label{eq:vgar-obj-before-jensen}
\end{align}
}%
\rk{where $q_{\phi}(\bm{Z} | \mathcal{G})$, parameterized by ${\phi}$, models the recognition network for approximate posterior inference.} It is given by
{\small
\begin{align}
	q_{\phi}(\bm{Z} | \mathcal{G})         & = \prod \limits_i^N q_{\phi}(\bm{z_i} | \mathcal{G}) \label{eq:vgae-qz_given_G} \\
	\quad q_{\phi}(\bm{z_i} | \mathcal{G}) & = \mathcal{N}\Big(\bm{\mu_i}(\mathcal{G)}, \mathrm{diag}(\bm{\sigma_i^2}(\mathcal{G}))\Big) 
	\label{eq:vgae-qz_i_given_G}
\end{align}
}%
\ua{where $\bm{\mu}_i(.)$ and $\bm{\sigma}^2_i(.)$ are learnt using neural networks} \rk{where $\bm{\mu}_i(.)$ and $\bm{\sigma}^2_i(.)$ are learnt using graph convolution networks (GCN) \cite{gcn}} and samples of $q_{\phi}(\bm{Z} | \mathcal{G})$ are obtained from mean and variance using the reparameterization trick \cite{vae}. 
%To avoid numerical overflows, we follow VAE and learn log variance instead of variance.

% Todo: Put it in the next section
% VGAE model uses graph convolution networks\textit{(GCN)} \cite{gcn} to learn $\bm{\mu}_i(.)$ and $\bm{\sigma}^2_i(.)$. GCN architecture focuses on the neighbors of the node under consideration rather than the whole graph $\mathcal{G}$. 

In order to ensure computational tractability, we use Jensen's Inequality \cite{jensen} to get ELBO bound of Eq.~\eqref{eq:vgar-obj-before-jensen}. i.e.
{\small
\begin{align}
	log\Big(p(\mathcal{G})\Big) & \geq \mathbb{E}_{\bm{Z} \sim q_{\phi}(\bm{Z} | \mathcal{G})} \Big\{ log\Big(\frac{p( \bm{Z}) p_{\theta}(\mathcal{G}| \bm{Z})}{q_{\phi}(\bm{Z} | \mathcal{G})}\Big)\Big\}                                                                                            \\
	                            & = \mathbb{E}_{\bm{Z} \sim q_{\phi}(\bm{Z} | \mathcal{G})} \Big\{ log\Big( p_{\theta}(\mathcal{G}| \bm{Z})\Big)\Big\} \nonumber \\ 
								& + \mathbb{E}_{\bm{Z} \sim q_{\phi}(\bm{Z} | \mathcal{G})} \Big\{ log\Big(\frac{p( \bm{Z}) }{q_{\phi}(\bm{Z} | \mathcal{G})}\Big)\Big\} \label{o-vgae} \\
								& = -\mathrm{BCE} - D_{KL}\Big(q_{\phi}(\bm{Z} | \mathcal{G}) || p(\bm{Z}) \Big) \label{eq:vgae-obj-abstract}
\end{align}
}%
% Todo: ask Rayyan about  A˜ = D− 1 2 AD− 1. 2 is the symmetrically normalized adjacency matri

\ua{where $p_{\theta}(\mathcal{G}| \bm{Z})$ corresponds to the decoder network, i.e. reconstructing the graph from the latent variables.
For an unweighted and undirected graph $\mathcal{G}$, we follow \cite{vgae} and restrict the decoder to reconstruct only edge information from the latent space.
The edge information can be represented by an adjacency matrix $\bm{A} \in \mathbb{R}^{N \times N}$ where $a_{ij}$ refers to the element in $i^{th}$ row and $j^{th}$ column. If an edge exists between node $i$ and $j$, we have $a_{ij} = 1$. 
Thus, the decoder is modeled by
{\small
\begin{align}
	  & p(\bm{A}| \bm{Z}) = \prod \limits_{(i,j)=(1,1)}^{(N,N)}p(a_{ij} = 1| \bm{z_i}, \bm{z_j}) \label{eq:pA-given-Z}  \\ 
	  & p(a_{ij} = 1| \bm{z_i}, \bm{z_j}) = \sigma(<\bm{z_i}, \bm{z_j}>) \nonumber,   
\end{align}
}%
where $<.\ ,\  .>$ denotes dot product and $\sigma(.)$ is the logistic sigmoid function. 
Hence, the Eq. \eqref{o-vgae} can be re-written as
{\small
\begin{align}
	log\Big(p(\mathcal{G})\Big) & = -\mathrm{BCE} - D_{KL}\Big(q(\bm{Z} | \mathcal{G}) || p(\bm{Z}) \Big) ,
\end{align}
}%
}
where $\mathrm{BCE}$ denotes binary cross-entropy loss between input edges and the reconstructed edges. 
$D_{KL}$ denotes the Kullback-Leibler (KL) divergence. \rk{By using \eqref{eq:vgae-pz}, \eqref{eq:vgae-pz_i}, \eqref{eq:vgae-qz_given_G} and \eqref{eq:vgae-qz_i_given_G}, the loss function of pure VGAE can be formulated as negative of \eqref{eq:vgae-obj-abstract} i.e.
{\small
\begin{align}
	L & = \mathrm{BCE} \nonumber \\
	  & + \sum \limits_{i=1}^{N}D_{KL}\bigg(\mathcal{N}\Big(\bm{\mu}_i(\mathcal{G}), \bm{\sigma}^2_i(\mathcal{G})\Big) \ || \  \mathcal{N}(\bm{0}, \mathrm{diag}(\bm{1})) \bigg) \label{eq:vgae-final-loss} 
\end{align}
}%
}
% todo check about KL-term
% ???We refer to the KL-divergence term of this loss as KL-term in the remainder of this paper.
\section{Over-pruning in pure VGAE} \label{sec:overpruning}
Burda et al. \cite{importance-weighted-autoencoders} showed that learning capacity of VAE is limited by \textit{over-pruning}. 
Several other studies \cite{over-pruning1,over-pruning2,over-pruning3,epitomic-vae}
confirm this and propose different remedies for the over-pruning problem.
They hold \ua{KL-term} \rk{the KL-divergence term} in the loss function of VAE responsible for over-pruning.
This term forces the latent variables to follow standard gaussian distribution. 
Consequently, those variables which fail to encode \textit{enough} information about input data are harshly penalized. 
In other words, if a latent variable is contributing little to the reconstruction, 
the variational loss is minimized easily by ``turning off" the corresponding hidden unit.
Subsequently, such variables simply collapse to the prior, i.e. generate standard gaussian noise.
We refer to the hidden units contributing to the reconstruction as \emph{active units} and the turned-off units as \emph{inactive units}.
The activity of a hidden unit $u$ was quantified by Burda et al. \cite{importance-weighted-autoencoders} via the statistic 
\begin{equation}
    A_u = Cov_x(\mathbb{E}_{u \sim q(u|x)}\{u\}).
    \label{eq:activity}
\end{equation}
A hidden unit $u$ is said to be \textit{active} if $A_u \geq 10^{-2}$.

\begin{figure}[t]
    \centering
	\begin{subfigure}{1\linewidth}
    \includegraphics[trim={0.5cm 1cm 1cm 1cm}, clip, width=0.95\linewidth]{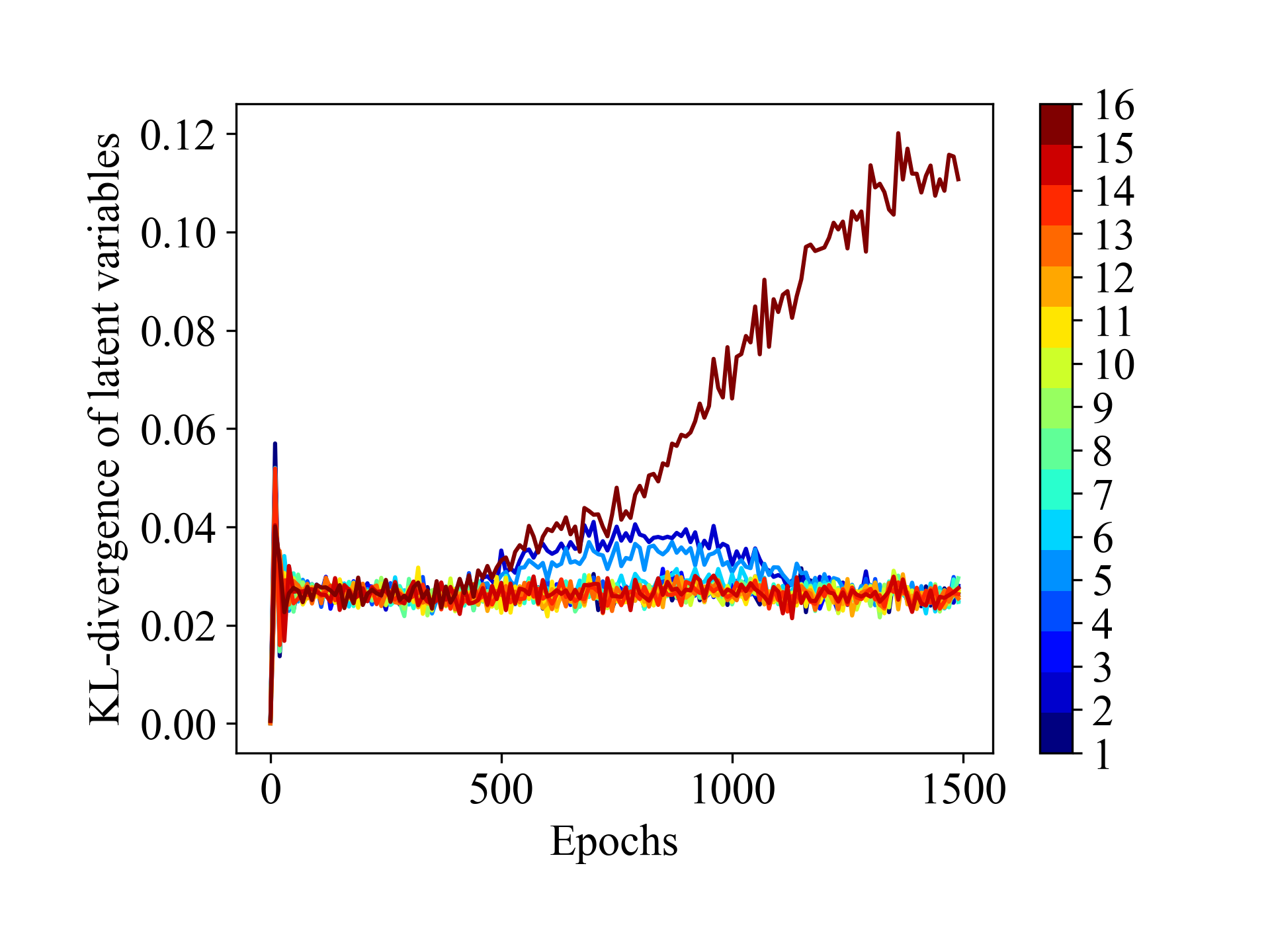}
	\caption{KL-divergence of latent variables in pure VGAE}
	\label{fig:vgae-normal-kld}
	\end{subfigure}
	\begin{subfigure}{1\linewidth}
    \includegraphics[trim={0.5cm 1cm 1cm 1cm}, clip, width=0.95\linewidth]{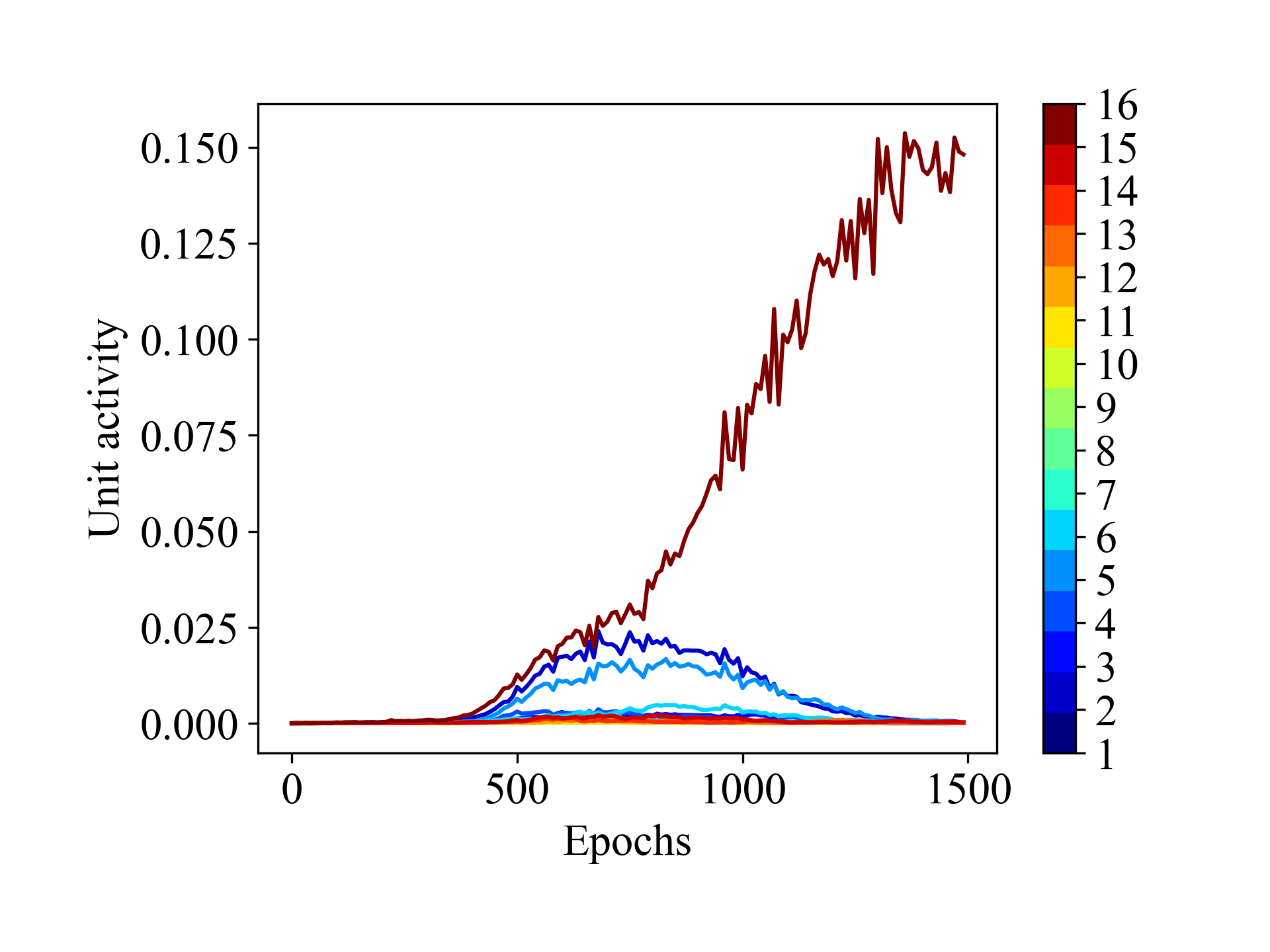}
	\caption{Unit activity of 16 hidden units in pure VGAE}
	\label{fig:vgae-normal-activity}
	\end{subfigure}
	\caption{\subref{fig:vgae-normal-kld} show that only one out of 16 hidden units is actively encoding input information required for the reconstruction. This is confirmed by the plot of unit activity in \subref{fig:vgae-normal-activity}.}
	\vspace{-5mm}
	\label{fig:vgae-normal}
\end{figure}

VGAE is an extension of VAE for graph data and loss functions of both models contain the \ua{KL-term} \rk{KL-divergence term}. 
Consequently, pure VGAE inherits the over-pruning issue. 
We verify this by training VGAE with Eq.~\eqref{eq:vgae-final-loss} on Cora dataset\footnote{Details of Cora dataset are given in experiments Sec.~\ref{sec: datasets}}.
We employ the same graph architecture as Kipf and Welling \cite{vgae}. The mean and log-variance of 16-dimensional latent space are learnt via Graph Convolutional Networks\cite{gcn}.
\ua{Fig.~\ref{fig:vgae-normal-logvar} shows the log variance of the
latent variables $\bm{z}$.  
We observe that 15 out of 16 latent variables have converged to a value around $-0.28$.
From Fig.~\ref{fig:vgae-normal-kld}, we observe that the KL-divergence of the latent variables exhibit a similar behavior. 
That is, only one latent variable seems to encode most of the information required for the reconstruction of the input.
}
\rk{From Fig.~\ref{fig:vgae-normal-kld}, we observe that 15 out of 16 latent variables have KL-divergence around $0.03$, indicating that they are very closely matched with standard gaussian distribution. Only one latent variable has managed to diverge in order to encode the information required by the decoder for reconstruction of the input.
}
% \begin{subfigure}{0.9\linewidth}
    % \includegraphics[trim={1cm 1cm 1.5cm 1cm},clip, width=0.95\linewidth]{res/logvar_vgae_beta_1}
	% \caption{Log variance of latent variables in pure VGAE}
	% \label{fig:vgae-normal-logvar}
	% \end{subfigure}

In other words pure VGAE model is using only one variable for encoding the input information while the rest 15 latent variables are not learning anything about the input.
These 15 latent variables collapse to the prior distribution and are simply generating standard gaussian noise. 
Fig.~\ref{fig:vgae-normal-activity} shows the activity of hidden units as defined in Eq.~\ref{eq:activity}.
It is clear that only one unit is \emph{active}, which corresponds to the latent variable with highest KL-divergence in the Fig.~\ref{fig:vgae-normal-kld}.
All other units have become \emph{inactive} and are not contributing in learning the reconstruction of the input.
This verifies the existence of \textit{over-pruning} in pure VGAE model.
\begin{figure}[t]
    \centering
	\begin{subfigure}{\linewidth}
    \includegraphics[trim={0.5cm 1cm 1cm 1cm}, clip, width=0.95\linewidth]{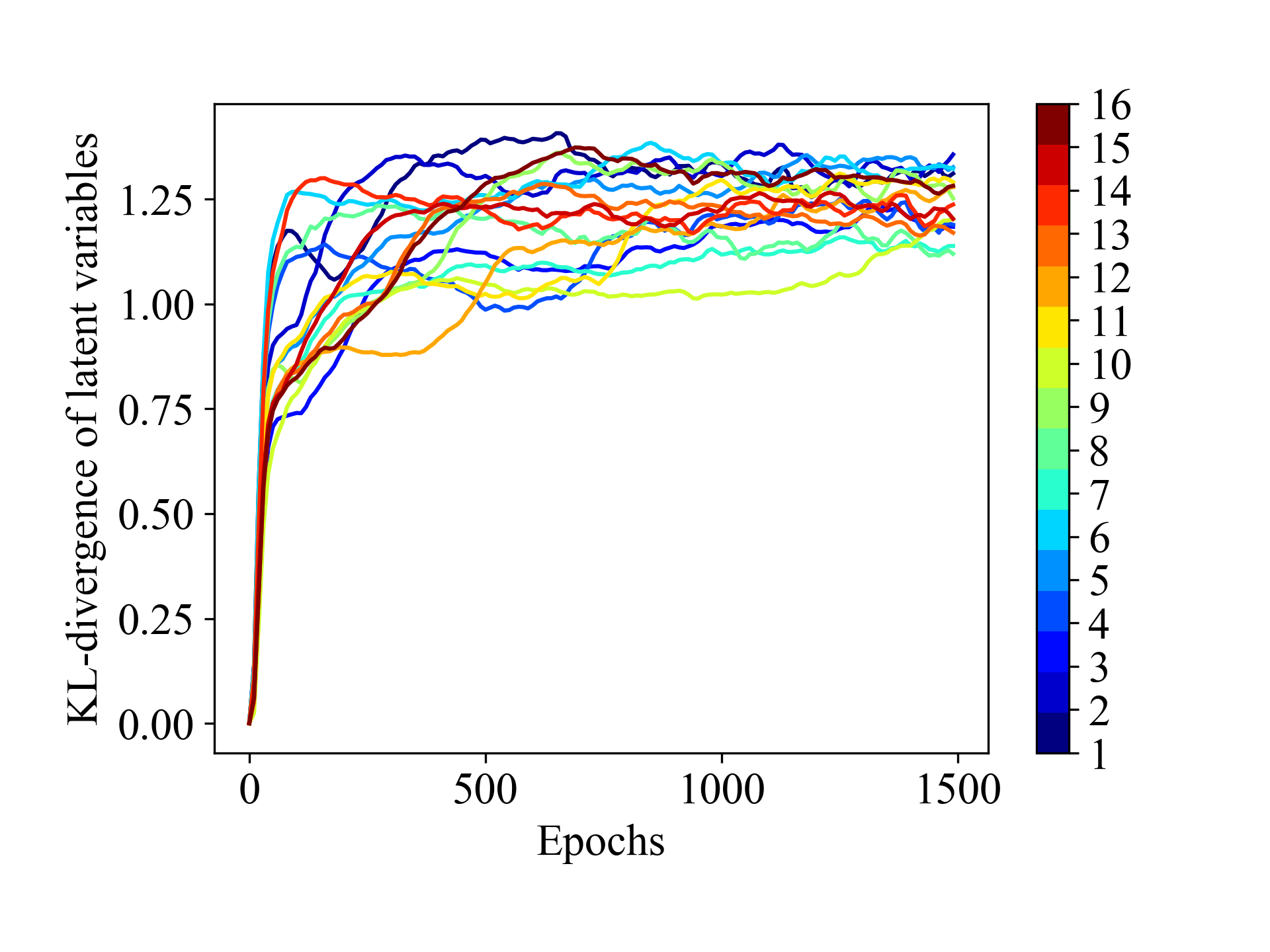}
	\caption{KL-divergence of latent variables: VGAE ($\beta$ \ua{=} \rk{$\approx$} 0.0003\cite{vgae})}
	\label{fig:vgae-beta-kld}
	\end{subfigure}
	\begin{subfigure}{\linewidth}
    \includegraphics[trim={0.5cm 1cm 1cm 1cm}, clip, width=0.95\linewidth]{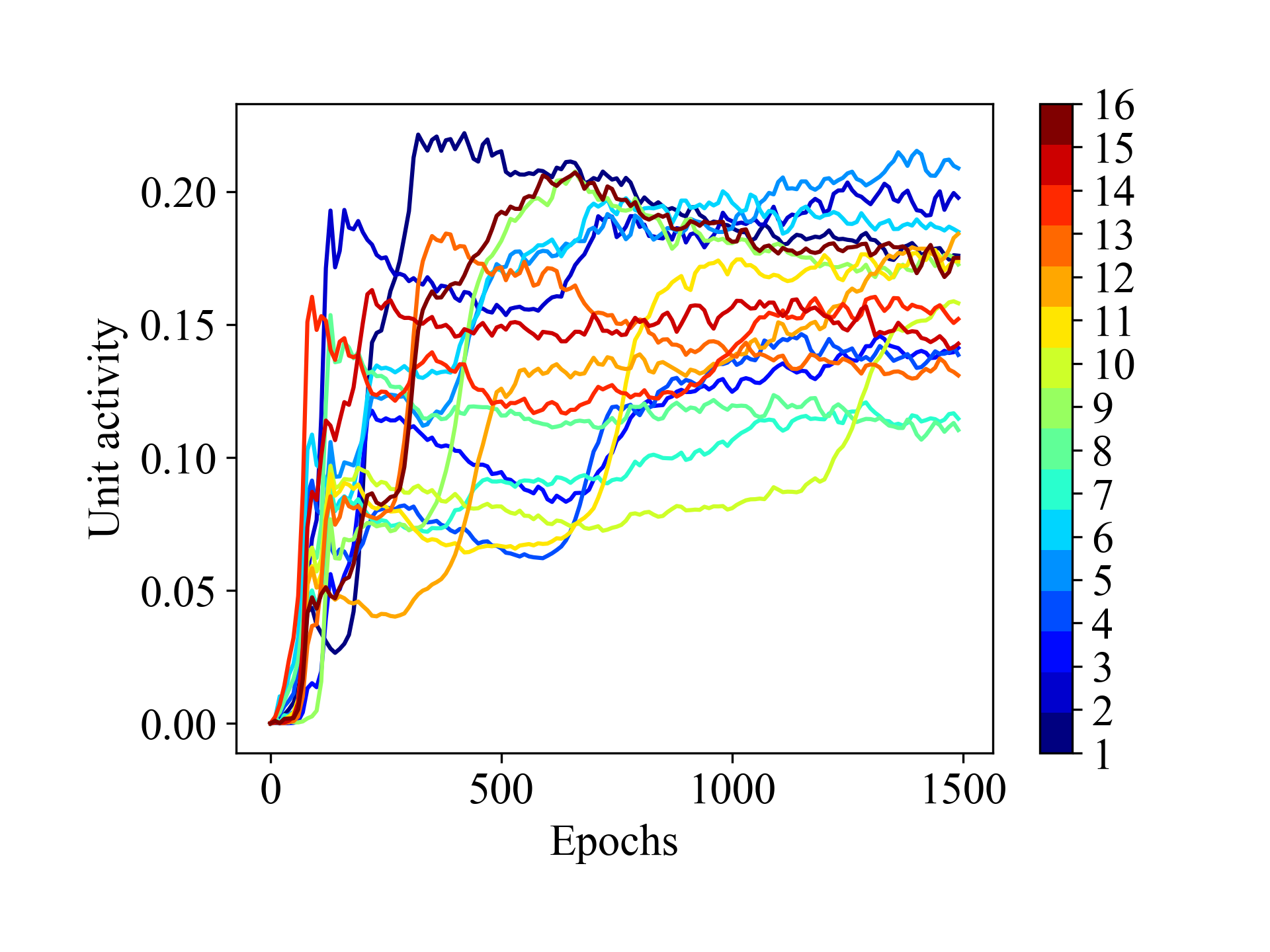}
	\caption{Unit activity of 16 hidden units: VGAE ($\beta$ \ua{=} \rk{$\approx$} 0.0003\cite{vgae})}
	\label{fig:vgae-beta-activity}
	\end{subfigure}
	\caption{All the hidden units are active but KL-divergence is quite \ua{low} \rk{high}, indicating poor matching of learnt distribution with prior, consequently  affecting generative ability of the model.}
	\label{fig:vgae-beta}
\end{figure}

\section{VGAE\cite{vgae}: Sacrificing Generative Ability for Handling Over-pruning} \label{sec:gen_ability}
Kipf and Welling's VGAE\cite{vgae} employed a simple way to get around the over-pruning problem by adding a penalty factor to the \ua{KL-term} \rk{KL-divergence} in Eq.~\eqref{eq:vgae-final-loss}. That is
{\small
\begin{equation}
	L = BCE + \beta \ D_{KL}\Big(q(\bm{Z} | \mathcal{G}) || p(\bm{Z}) \Big) .
% 	log\Big(p(\mathcal{G})\Big) = \mathbb{E}_{\bm{Z} \sim q(\bm{Z} | \mathcal{G})} \Big\{ log\Big( p(\mathcal{G}| \bm{Z})\Big)\Big\} + \beta \mathbb{E}_{\bm{Z} \sim q(\bm{Z} | \mathcal{G})} \Big\{ log\Big(\frac{p( \bm{Z}) }{q(\bm{Z} | \mathcal{G})}\Big)\Big\} 
	\label{loss-vgae-weighted}
\end{equation}
}%
But a consequence of using the penalty factor $\beta$ is poor generative ability of VGAE. 
We verify this by training VGAE on Cora dataset with varying $\beta$ in Eq.~\eqref{loss-vgae-weighted}. 
We call the penalty factor $\beta$, as the loss of $\beta$VAE ( \cite{betavae1,betavae2} ) has the same factor multiplied with its \ua{KL-term} \rk{KL-divergence term}.
Specifically, in $\beta$VAE, $\beta > 1$ is chosen to enforce better distribution matching. 
Conversely, a smaller $\beta$ is selected for relaxing the distribution matching, i.e. the latent distribution is allowed to be more different than the prior distribution.
This enables latent variables to learn better reconstruction at the expense of the generative ability.
In the degenerate case, when $\beta = 0$, VGAE model is reduced to non-variational graph autoencoder (GAE). 
VGAE as proposed by Kipf and Welling\cite{vgae} has the loss function similar to $\beta$VAE with $\beta$ chosen as reciprocal of number of nodes in the graph. As a result $\beta$ is quite small i.e. $\sim$ 0.0001-0.00001. 

Fig.~\ref{fig:vgae-beta} shows the KL-divergence and hidden unit activity for original VGAE\cite{vgae} model.
We observe that all the hidden units are active, i.e. $A_u \geq 10^{-2}$.
However, the value of KL-divergence is quite \ua{low} \rk{high} for all latent variables, indicating poor matching of $q_{\phi}(\bm{Z}|\mathcal{G})$ with the prior distribution. 
This adversely affects the generative ability of the model. 
Concretely, the variational model is supposed to learn such a latent representation which follows standard gaussian (prior) distribution.
Such \ua{low} \rk{high} values of KL-divergence implies that the learnt distribution is not standard gaussian. 
The reason is that \ua{KL-term} \rk{the KL-divergence term in \eqref{loss-vgae-weighted}} was responsible for ensuring that the posterior distribution being learned follows standard gaussian distribution. VGAE\cite{vgae} model assigns too small weight ($\beta$ = 0.0003) to the \ua{KL-term} \rk{KL-divergence term}.
Consequently, when new samples are generated from standard gaussian distribution $p(\bm{Z})$ and then passed through the decoder $p_{\theta}(\bm{A}|\bm{Z})$, we get quite different output than the graph data used for training. 

\begin{figure}[t]
    \centering
	\begin{subfigure}{\linewidth}
    \includegraphics[width=0.95\linewidth]{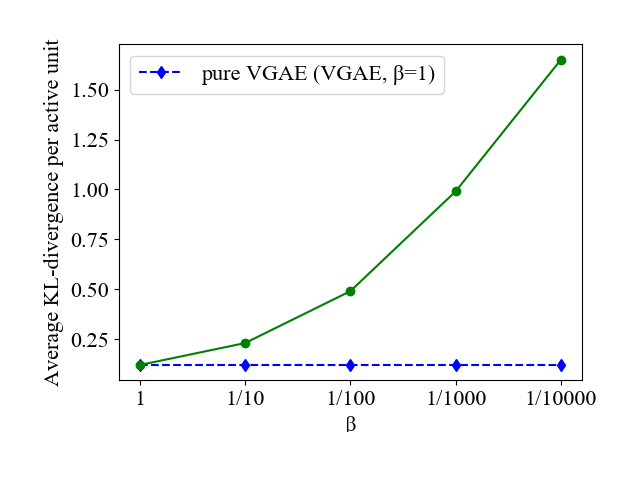}
	\caption{Change in the active units of original VGAE\cite{vgae}}
	\label{fig:beta-kld}
	\end{subfigure}
	\begin{subfigure}{\linewidth}
    \includegraphics[width=0.95\linewidth]{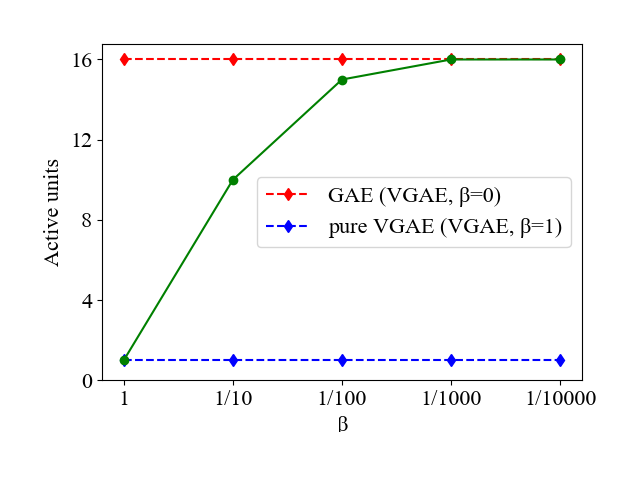}
	\caption{Change in the Average KL-divergence per active unit }
	\label{fig:beta-activity}
	\end{subfigure}
	\caption{Effect of varying $\beta$ on original VGAE\cite{vgae}} 
	\label{fig:beta}
\end{figure}

Fig.~\ref{fig:beta} shows that Kipf and Welling's\cite{vgae} approach to deal with over-pruning makes VAGE similar to its non-variational counter-part i.e. graph autoencoder (GAE).
As $\beta$ is decreased, VGAE model learns to give up on the generative ability and behaves similar to GAE. 
This can be seen in Fig.~\ref{fig:beta} (a), where the average KL-divergence per active hidden unit \ua{decreases} \rk{increases} drastically as $\beta$ becomes smaller.
On the other hand, we observe from Fig.~\ref{fig:beta} (b) that decreasing $\beta$ results in higher number of active hidden units till it achieves the same number as GAE.

We conclude that as the contribution of \ua{KL-term} \rk{KL-divergence} is penalized in the loss function (Eq.~\ref{loss-vgae-weighted}), VGAE model learns to sacrifice the generative ability for avoiding over-pruning. Conversely, VGAE handles the over-pruning problem by behaving like a non-variational model GAE \cite{vgae-to-gae-github-comment}.

% This issue can be solved by:
% \begin{enumerate}
%     \item Adding a small weight to second term of \ref{o-vae} so that there is less stress on the model to match $q(z|x)$ with the prior. However, this will result in a poorly learnt latent distribution and affect generative ability of the network.
%     \item Adding dropout. However this will just add redundancy to the model.
%     \item Using a model based approach to force a set of   
% \end{enumerate}
	
% \begin{itemize}
%     \item pruning happens in vae too
%     \item epitomic vae for normal datasets
%     \item s-vgae for graph datasets
% \end{itemize}

\section{Epitomic Variational Graph Autoencoder}
We propose epitomic variational graph autoencoder (EVGAE) 
which generalizes and improves the VGAE model. 
EVGAE not only successfully mitigates the over-pruning issue of pure VGAE but also attains better generative ability than VGAE\cite{vgae}. 
The motivation comes from the observation that for a certain graph node, a subset of the latent variables suffices to yield good reconstruction of edges. 
Yeung et al. \cite{epitomic-vae} proposed a similar solution for tackling over-pruning problem in VAE.
We assume $M$ subsets of the latent variables called \textit{epitomes}. 
They are denoted by $\{\mathcal{D}_1, \cdots, \mathcal{D}_M\}$. 
Furthermore, it is ensured that every subset shares some latent variables with at least one other subset. 
% TODO:??? Phrase this penalization point accurately. in reference to loss terms if possible.
We penalize only one epitome for an input node. 
% As a result, 
This encourages other epitomes to be active. 
Let $y_i$ denote a discrete random variable that decides which epitome is penalized for a node $i$. 
For a given node, the prior distribution of $y_i$ is assumed to be uniform over all the epitomes.
$\bm{y}$ represents the stacked random vector for all $N$ nodes of the graph. So:

{\small
\begin{equation}
	p(\bm{y}) = \prod \limits_{i=0}^{N} p(y_i); \quad p(y_i) = \mathcal{U}(1, M),
\end{equation}
}%
where $\mathcal{U}(\cdot)$ denotes uniform distribution.

\begin{figure}
	\centering
	\begin{tikzpicture}[scale=0.5]
		\def\BITARRAY{
			{1,1,1,0,0,0,0,0,0,0,0,0,0,0,0,0},
			{0,0,1,1,1,0,0,0,0,0,0,0,0,0,0,0},
			{0,0,0,0,1,1,1,0,0,0,0,0,0,0,0,0},
			{0,0,0,0,0,0,1,1,1,0,0,0,0,0,0,0},
			{0,0,0,0,0,0,0,0,1,1,1,0,0,0,0,0},
			{0,0,0,0,0,0,0,0,0,0,1,1,1,0,0,0},
			{0,0,0,0,0,0,0,0,0,0,0,0,1,1,1,0},
			{1,0,0,0,0,0,0,0,0,0,0,0,0,0,1,1}%
		}
		\fill[gray]
		\foreach \row [count=\y] in \BITARRAY {
			\foreach \cell [count=\x] in \row {
				\ifnum\cell=1 %
					(\x-1, -\y+1) rectangle ++(1, -1)
				\fi
				\pgfextra{%
					\global\let\maxx\x
					\global\let\maxy\y
				}%
			}
		}
		;
		\draw[thin] (0, 0) grid[step=1] (\maxx, -\maxy);
	\end{tikzpicture}
	\caption{Example of eight epitomes in a 16-dimensional latent space.}
	\vspace{-5mm}
	\label{epitomes-example}
\end{figure}
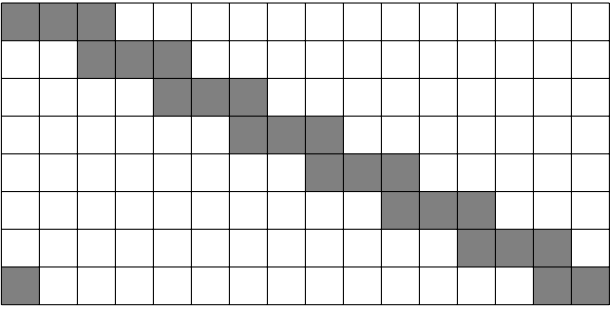

Let $\bm{E} \in \mathbb{R}^{M \times D}$ denote a binary matrix, where each row 
represents an epitome and each column represents a latent variable. 
Fig.~\ref{epitomes-example} shows $\bm{E}$ with $M = 8$ and $D = 16$ in a $D$-dimensional latent space. The grayed squares of $r^{th}$ row show the latent variables which constitute the epitome $\mathcal{D}_r$. 
We denote $r^{th}$ row of $\bm{E}$ by %$\bm{E}_r$. % $\bm{e}_r$
$\bm{E}[r, :]$. 
% For instance, in the Fig. \ref{epitomes-example}, $\bm{E}[1, :] = [0, 0, 1, 1, 1, 0, 0, 0, 0, 0, 0, 0, 0, 0, 0, 0]$.

\ua{\subsubsection{Generative Model} of EVGAE is the same as that of VGAE i.e.:
{\small
\begin{equation}
	p(\mathcal{G}, \bm{Z}) = p(\bm{Z}) p_{\theta}(\mathcal{G} | \bm{Z})
\end{equation}
}%
}
\ff{
\subsubsection{Generative Model} of EVGAE is given by:
{\small
\begin{equation}
	p(\mathcal{G}, \bm{Z}, \bm{y}) = p(\bm{y}) p(\bm{Z} | \bm{y}) p_{\theta}(\mathcal{G} | \bm{Z}),
\end{equation}
}%
where
{\small
\begin{align}
    p(\bm{Z} | \bm{y}) & = \prod \limits_{i=0}^{N} p(\bm{z}_i | y_i) \\
    p(\bm{z}_i | y_i) & = \prod \limits_{j=1}^{D}\Big( E[y_i, j] \ \mathcal{N}(0, 1) + (1 - E[y_i, j]) \delta(0) \Big), \label{eq:pz_i_given_y_i}
\end{align}
where $\bm{E}[y_i, j]$ refers to $j^{th}$ component of epitome $y_i$. Eq.~\eqref{eq:pz_i_given_y_i} shows that $\bm{z}_i | y_i$ follows standard gaussian distribution for the latent variables $j$ where $E[y_i, j] = 1$ and for the rest it follows degenerate distribution $\delta(0)$ located at $0$. 
}%

}
% Where:
% \begin{align}
% 	p(\bm{Z}) & = \sum \limits_y p(\bm{y}) p(\bm{Z} | \bm{y}) \nonumber \\
% 	p(\bm{Z} | \bm{y}) & = \prod \limits_{i=1}^{N} p(\bm{z}_i | y_i); \quad p(\bm{z}_i |y_i) = \prod \limits_{j=1}^{D} p(z_{ij} | y_i)\nonumber \\
% 	p(z_{ij} | y_i)    & = \begin{cases}                                                                                                        
% 	\mathcal{N}(0, 1)  & if \quad \bm{E}[y_i,j]=1                                                                                               \\
% 	\delta(0)          & otherwise                                                                                                              
% 	\end{cases}
% \end{align}
% % \mathcal{N}\Big(\bm{0}, \big(\bm{E}[y, :] \odot diag(\bm{1})\big) + \big(\big(\bm{1} - \bm{E}[y, :]\big) \odot diag(\bm{\gamma})\big)\Big) 
% i.e. the components active for epitome $y_i$ have unit variance and the other components are collapsed to dirac delta function. Like in the case of VGAE, we limit ourselves to recover edge information from latent space where the edges are unweighted and undirected. So the decoder is the same as in Eq. \ref{eq:pA-given-Z}

\subsubsection{Inference Model} uses the following approximate posterior:
{\small
\begin{align}
	&q_{\phi}(\bm{Z}, \bm{y} | \mathcal{G})  = q_{\phi}(\bm{y} | \mathcal{G})  q_{\phi}(\bm{Z} | \mathcal{G}),
\end{align}
}%
with
{\small
\begin{align}
	&q_{\phi}(\bm{y} | \mathcal{G})          = \prod \limits_{i=1}^{N} q_{\phi}(y_i | \mathcal{G}) \\
	&q_{\phi}(y_i | \mathcal{G}) = \mathrm{Cat}(\bm{\pi}_i(\mathcal{G}))                                                                  \\
	&q_{\phi}(\bm{Z} | \mathcal{G})          = \prod \limits_i^N q_{\phi}(\bm{z}_i | \mathcal{G}) \\
	& q_{\phi}(\bm{z}_i | \mathcal{G}) = \mathcal{N}\Big(\bm{\mu}_i(\mathcal{G)}, \mathrm{diag}(\bm{\sigma}_i^2(\mathcal{G}))\Big) ,
\end{align}
}%

where $\mathrm{Cat}(.)$ refers to the categorical distribution. $\bm{\pi}_i(.)$, 
$\bm{\mu}_i(.)$ and $\bm{\sigma}_i^2(.)$ are learnt using \ua{separate multi-layer perceptron (MLP)} \rk{two-layer GCN networks}.
\ff{Specifically, $\bm{\pi}_i(.)$ is obtained by learning a real vector which is then passed through softmax layer to give probabilities.}
Under the assumption that $\bm{y}$ and ${\mathcal{G}}$ are independent, given $\bm{Z}$; the objective function is given by
{\footnotesize
\begin{align}
	log\Big(p(\mathcal{G})\Big) & = log\Big(\int{ \sum \limits_{\bm{y}} p(\bm{y}) p( \bm{Z} | \bm{y}) p_{\theta}(\mathcal{G}| \bm{Z}) \ d\bm{Z}}\Big) \label{eq:generic-evgae-obj}                                                                 \\
	                            & = log\Big(\mathbb{E}_{(\bm{Z}, \bm{y}) \sim q_{\phi}(\bm{Z}, \bm{y} | \mathcal{G})} \Big\{\frac{p(\bm{y})p( \bm{Z} | \bm{y}) p_{\theta}(\mathcal{G}| \bm{Z})}{q_{\phi}(\bm{Z},  \bm{y}| \mathcal{G})}\Big\}\Big)               \\                                
	                            & = log\Big(\mathbb{E}_{(\bm{Z}, \bm{y}) \sim q_{\phi}(\bm{Z}, \bm{y} | \mathcal{G})} \Big\{\frac{p(\bm{y})p( \bm{Z} | \bm{y}) p_{\theta}(\mathcal{G}| \bm{Z})}{q_{\phi}(\bm{Z}| \mathcal{G}) q_{\phi}(\bm{y}| \mathcal{G})}\Big\}\Big) .
\end{align}
}%
By using Jensen's inequality \cite{jensen}, the ELBO bound for log probability becomes
{\small
\begin{align}
	log\Big(p(\mathcal{G})\Big) & \geq \mathbb{E}_{(\bm{Z}, \bm{y}) \sim q_{\phi}(\bm{Z}, \bm{y} | \mathcal{G})} \Big\{ log\Big(\frac{p(\bm{y})p( \bm{Z} | \bm{y}) p_{\theta}(\mathcal{G}| \bm{Z})}{q_{\phi}(\bm{Z} | \mathcal{G}) q_{\phi}(\bm{y}| \mathcal{G})}\Big)\Big\} \\
	                            & = \mathbb{E}_{\bm{Z} \sim q_{\phi}(\bm{Z} | \mathcal{G})} \Big\{ log\Big( p_{\theta}(\mathcal{G}| \bm{Z})\Big)\Big\} \nonumber                                                                                               \\
	                            & + \mathbb{E}_{\bm{y} \sim q_{\phi}(\bm{y} | \mathcal{G})} \Big\{ log\Big(\frac{p(\bm{y}) }{q_{\phi}(\bm{y} | \mathcal{G})}\Big)\Big\} \nonumber                                                                            \\
	                            & + \mathbb{E}_{(\bm{Z}, \bm{y}) \sim q_{\phi}(\bm{Z}, \bm{y} | \mathcal{G})} \Big\{ log\Big(\frac{p( \bm{Z} | \bm{y}) }{q_{\phi}(\bm{Z} | \mathcal{G})}\Big)\Big\}. \label{o-evgae}                                          
\end{align}
}%

Following VGAE\cite{vgae}, we restrict the decoder to recover only edge information from the latent space.
% where the edges are unweighted and undirected. 
Hence, the decoder is the same as in Eq.~\eqref{eq:pA-given-Z}. 
Thus, the first term in Eq.~\eqref{o-evgae} simplifies in a similar way as in VGAE i.e. binary cross-entropy between input and reconstructed edges. 

\begin{table*}[ht]
	\centering
	\caption{Results of link prediction on citation datasets}						
	\resizebox{\textwidth}{!}{%
		\begin{tabular}{p{2.5cm}|c|c|c|c|c|c}
			\hline
			\hline
			\multirow{2}{*}{\textbf{Method}} & \multicolumn{2}{c}{\textbf{Cora}} & \multicolumn{2}{c}{\textbf{Citeseer}} & \multicolumn{2}{c}{\textbf{PubMed}} \\
			                    & AUC                               & AP                                & AUC                               & AP                                & AUC                               & AP                                \\
			\hline
			\hline
			DeepWalk            & $83.1 \pm 0.01$                   & $85.0 \pm 0.00$                   & $80.5 \pm 0.02$                   & $83.6 \pm 0.01$                   & $84.4 \pm 0.00$                   & $84.1 \pm 0.0$                    \\
			\hline
			Spectral Clustering & {$84.6 \pm 0.01$}  & {$88.5 \pm 0.00$}  & {$80.5 \pm 0.01$}  & {$85.0 \pm 0.01$}  & {$84.2 \pm 0.02$}  & {$87.8 \pm 0.01$}  \\
			\hline
			GAE (VGAE\cite{vgae} with $\beta$ = 0)                & \multirow{2}{*}{$91.0 \pm 0.02$}                   & \multirow{2}{*}{$92.0 \pm 0.03$}                  & \multirow{2}{*}{$89.5 \pm 0.04$ }                  & \multirow{2}{*}{$89.9 \pm 0.05$}                   & \multirow{2}{*}{$96.4 \pm 0.00$}                   & \multirow{2}{*}{$96.5 \pm 0.0$}                    \\
			\hline
			VGAE \cite{vgae} ($\beta$ $\sim 10^{-4}-10^{-5}$)              & \multirow{2}{*}{$91.4 \pm 0.01$ }                  & \multirow{2}{*}{$92.6 \pm 0.01$  }                 & \multirow{2}{*}{$90.8 \pm 0.02$ }                  & \multirow{2}{*}{$92.0 \pm 0.02$}                   & \multirow{2}{*}{$94.4 \pm 0.02$    }               & \multirow{2}{*}{$94.7 \pm 0.0$}                    \\
			\hline
			pure VGAE ($\beta$ = 1)    & {$79.44 \pm 0.03$} & {$80.51 \pm 0.02$} & {$77.08 \pm 0.03$} & {$79.07 \pm 0.02$} & {$82.79 \pm 0.01$} & {$83.88 \pm 0.01$} \\
			\hline
			EVGAE   ($\beta$ = 1)             & $\bm{92.96 \pm 0.02}$             & $\bm{93.58 \pm 0.03}$             & $\bm{91.55 \pm 0.03}$             & $\bm{93.24 \pm 0.02}$             & $\bm{96.80 \pm 0.01}$             & $\bm{96.91 \pm 0.02}$             \\
			\hline
		\end{tabular}%
	}
	\label{tab:link-prediction-res}
\end{table*}

The second term in Eq.~\eqref{o-evgae} is computed as:
{\small
\begin{align}
	\mathbb{E}_{\bm{y} \sim q_{\phi}(\bm{y} | \mathcal{G})} &\Big\{ log\Big(\frac{p(\bm{y}) }{q_{\phi}(\bm{y} | \mathcal{G})}\Big)\Big\}  = \mathbb{E}_{\bm{y} \sim q_{\phi}(\bm{y} | \mathcal{G})} \Big\{\sum \limits_{i=1}^{N} log\Big(\frac{p(y_i) }{q_{\phi}(y_i | \mathcal{G})}\Big)\Big\}\nonumber \\
	                                                                                                                      & = \sum \limits_{i=1}^{N} \ \mathbb{E}_{y_i \sim q_{\phi}(y_i | \mathcal{G})} \Big\{ log\Big(\frac{p(y_i) }{q_{\phi}(y_i | \mathcal{G})}\Big)\Big\} \nonumber   \\
	                                                                                                                      & = - \sum \limits_{i=1}^{N} \ D_{KL} \Big(q_{\phi}(y_i | \mathcal{G}) || p(y_i)\Big) \nonumber                                                           \\
	                                                                                                                      & = - \sum \limits_{i=1}^{N} \ D_{KL} \Big(\mathrm{Cat}(\bm{\pi}_i(\mathcal{G})) || \  \mathcal{U}(1, M)\Big).                                                    
\end{align} 
}%
% & = - \sum \limits_{i=1}^{N} \ D_{KL} \Big(\mathcal{N}(\bm{\mu}_i(\mathcal{G}, diag(\bm{\sigma^2(\mathcal{G})}))) || p(y_i)\Big)
The third term in Eq.~\eqref{o-evgae} is computed as follows:
{\small
\begin{align}
	  & \mathbb{E}_{(\bm{Z}, \bm{y}) \sim q_{\phi}(\bm{Z}, \bm{y} | \mathcal{G})} \Big\{ log\Big(\frac{p( \bm{Z} | \bm{y}) }{q_{\phi}(\bm{Z}| \mathcal{G})}\Big)\Big\} \nonumber                                                           \\
	= & \mathbb{E}_{\bm{y} \sim q_{\phi}(\bm{y} | \mathcal{G})} \bigg\{ \mathbb{E}_{\bm{Z} \sim q_{\phi}(\bm{Z} | \mathcal{G})} \Big\{ log\Big(\frac{p( \bm{Z} | \bm{y}) }{q_{\phi}(\bm{Z} | \mathcal{G})}\Big) \Big\}\bigg\} \nonumber           \\
	= & \sum \limits_{\bm{y}} q_{\phi}(\bm{y} | \mathcal{G}) \mathbb{E}_{\bm{Z} \sim q_{\phi}(\bm{Z} | \mathcal{G})} \Big\{ log\Big(\frac{p( \bm{Z} | \bm{y}) }{q_{\phi}(\bm{Z} | \mathcal{G})}\Big) \Big\} \nonumber                             \\
	= & \sum \limits_{i=1}^{N} \sum \limits_{\bm{y}} q_{\phi}(\bm{y} | \mathcal{G}) \mathbb{E}_{\bm{z}_i \sim q_{\phi}(\bm{z}_i | \mathcal{G})} \Big\{ log\Big(\frac{p( \bm{z}_i | y_i) }{q_{\phi}(\bm{z}_i | \mathcal{G})}\Big) \Big\} \nonumber \\
	= & \sum \limits_{i=1}^{N} \sum \limits_{y_i} q_{\phi}(y_i | \mathcal{G}) \mathbb{E}_{\bm{z}_i \sim q_{\phi}(\bm{z}_i | \mathcal{G})} \Big\{ log\Big(\frac{p( \bm{z}_i | y_i) }{q_{\phi}(\bm{z}_i | \mathcal{G})}\Big) \Big\}                 \nonumber \\
	= & - \sum \limits_{i=1}^{N} \sum \limits_{y_i} q_{\phi}(y_i | \mathcal{G}) D_{KL} \Big( q_{\phi}(\bm{z}_i | \mathcal{G}) || p( \bm{z}_i | y_i) \Big)
	\label{eq:kl-evgae-intermediate}
\end{align}
}%
We take motivation from \cite{epitomic-vae} to compute Eq.~\eqref{eq:kl-evgae-intermediate} as:
{\small
\begin{align}
	  & - \sum \limits_{i=1}^{N} \sum \limits_{y_i} q_{\phi}(y_i | \mathcal{G}) D_{KL} \Big( q_{\phi}(\bm{z}_i | \mathcal{G}) || p( \bm{z}_i | y_i) \Big) \nonumber                             \\
	= & - \sum \limits_{i=1}^{N} \sum \limits_{y_i} q_{\phi}(y_i | \mathcal{G}) \sum \limits_{j = 1}^{D}  \bm{E}[y_i, j] D_{KL} \Big( q_{\phi}(z_i^j | \mathcal{G}) || p( z_i^j) \Big) \\
	= & - \sum \limits_{i=1}^{N} \sum \limits_{y_i} \bm{\pi}_i(\mathcal{G}) \sum \limits_{j = 1}^{D} \bm{E}[y_i, j] \nonumber \\
	& \quad \quad \quad \quad \quad \quad D_{KL} \bigg( \mathcal{N}\Big(\mu_i^j(\mathcal{G}), (\sigma_i^2)^j(\mathcal{G})\Big) || \mathcal{N}(0, 1)\bigg),
	\label{eq:kl-evgae}
\end{align}
}%

% The KL-divergence term in above equation is computed as
% {\small
% \begin{align}
% 	D_{KL} \Big( q(z_i^j | \mathcal{G}) || p( z_i^j) \Big) =  D_{KL} \bigg( \mathcal{N}\Big(\mu_i^j(\mathcal{G}), (\sigma_i^2)^j(\mathcal{G})\Big) || \mathcal{N}(0, 1)\bigg).
% 	\label{eq:kl-evgae}
% \end{align}
% }%

% {\small
% \begin{align}
% 	- \sum \limits_{i=1}^{N} \sum \limits_{y_i} \bm{\pi}_i(\mathcal{G}) \sum \limits_{j = 1}^{D} & \bm{E}[y_i, j] . \nonumber \\
% 	& D_{KL} \bigg( \mathcal{N}\Big(\mu_i^j(\mathcal{G}), (\sigma_i^2)^j(\mathcal{G})\Big) || \mathcal{N}(0, 1)\bigg).
% 	\label{eq:kl-evgae}
% \end{align}
% }%

where $z_i^j$ denotes $j^{th}$ component of vector $\bm{z}_i$. In Eq.~\eqref{eq:kl-evgae}, for each node, we sum over all the epitomes. 
For a given epitome, we only consider the effect of those latent variables which are selected by $\bm{E}$ for that epitome. This also implies that the remaining latent variables have the freedom to better learn the reconstruction. 
Consequently, EVGAE encourages more hidden units to be active without penalizing the hidden units which are contributing little to the reconstruction.
The final loss function is given by:
{\small
\begin{align}
	L & = \mathrm{BCE} + \sum \limits_{i=1}^{N} \ D_{KL} \Big(\mathrm{Cat}(\bm{\pi}_i(\mathcal{G})) || \  \mathcal{U}(1, M)\Big)\nonumber \\
	& + \sum \limits_{i=1}^{N} \sum \limits_{y_i} \bm{\pi}_i(\mathcal{G}) \sum \limits_{j = 1}^{D} \bm{E}[y_i, j] \nonumber \\
	& \quad \quad \quad \quad \quad \quad D_{KL} \bigg( \mathcal{N}\Big(\mu_i^j(\mathcal{G}), (\sigma_i^2)^j(\mathcal{G})\Big) || \mathcal{N}(0, 1)\bigg).
\label{eq:evgae-final-loss} 
\end{align}
}%

VGAE model can be recovered from EVGAE model, if we have only one epitome consisting of all latent variables.
Hence the model generalizes VGAE. 
The algorithm for training EVGAE is given in Algo.~\ref{alg:evgae}.

\begin{algorithm}[t]
	\SetAlgoLined
	\textbf{Input: } \begin{itemize}
	\item $\mathcal{G}$
	\item Epochs
	\item The matrix $\bm{E}$ to select latent variables \\ for each epitome.
	\end{itemize}
																								
	% \KwResult{Write here the result }
	Initialize model weights;
	$i=1$
																								
	\While{$e \leq Epochs$}{
		compute $\bm{\pi_i(.)}$, $\bm{\mu_i(.)}$ and $\bm{\sigma_i^2(.)} \ \forall i$\;
		compute $\bm{z}_i \ \forall i$ by reparameterization trick\;
		compute loss using Eq. \eqref{eq:evgae-final-loss}\;
		update model weights using back propagation
	}
	\caption{EVGAE Algorithm}
	\label{alg:evgae}
\end{algorithm}

\section{Experiments}

\subsection{Datasets} \label{sec: datasets}
We compare the performance of EVGAE with several baseline methods on the link prediction task.
We conduct the experiments on three benchmark citation datasets\cite{citation-datasets}.

\textbf{Cora} dataset has 2,708 nodes with 5,297 undirected and unweighted links. The nodes are defined by 1433 dimensional binary feature vectors, divided in 7 classes.

\textbf{Citeseer} dataset has 3,312 nodes defined by 3703 dimensional feature vectors. The nodes are divided in 6 distinct classes. There are 4,732 links between the nodes.

\textbf{PubMed} consists of 19,717 nodes defined by 500 dimensional feature vectors linked by 44,338 unweighted and undirected edges. These nodes are divided in 3 classes.

\subsection{Implementation Details and Performance Comparison}
In order to ensure fair comparison, we follow the experimental setup of Kipf and Welling\cite{vgae}.
That is, we train the EVGAE and pure VGAE model on an incomplete version of citation datasets. Concretely, the edges of the dataset are divided in training set, validation set and test set. 
Following \cite{vgae}, we use 85\% edges for training, 5\% for validation and 10\% for testing the performance of the model. 

We compare the performance of EVGAE with three strong baselines, namely:  VGAE\cite{vgae}, spectral clustering\cite{spectral-clustering} and DeepWalk\cite{DW}. 
We also report the performance of pure VGAE ($\beta$=1) and GAE (VGAE with $\beta$=0). 
Since DeepWalk and spectral clustering do not employ node features; so VGAE, GAE and EVGAE have an undue advantage over them.
The implementation of spectral clustering is taken from \cite{spectral-clustering-impl} with 128 dimensional embedding and 
for DeepWalk, the standard implementation is used \cite{DW-impl}. 
For VGAE and GAE, we use the implementation provided by Kipf and Welling\cite{vgae}. 
EVGAE also follows a similar structure with latent embedding being 512 dimensional and the hidden layer consisting of 1024 hidden units, half of which learn $\bm{\mu}_i(.)$ and the other half for learning log-variance. 
We select 256 epitomes for all three datasets. 
Each epitome enforces three units to be active, while sharing one unit with neighboring epitomes. 
This can also be viewed as an extension of the matrix shown in Fig. \ref{epitomes-example}. 
Adam \cite{adam} is used as optimizer with learning rate $1e^{-3}$. 
Further implementation details of EVGAE can be found in the code \cite{evgae-github}.
% TODO: Anonymize it, ask me on how to do it via github.

For evaluation, we follow the same protocols as other recent works \cite{vgae}\cite{DW}\cite{spectral-clustering}. 
That is, we measure the performance of models in terms of area under the ROC curve (AUC) and average precision (AP) scores on the test set. 
We repeat each experiment 10 times in order to estimate the mean and the standard deviation in the performance of the models.

We can observe from Table~\ref{tab:link-prediction-res} that the results of EVGAE are competitive or slightly better than other methods. We also note that the performance of variational method pure VGAE is quite bad as compared to our variational method EVGAE. Moreover, the performance of methods with no or poor generative ability (GAE and VGAE \cite{vgae} with $\beta$ $\sim 10^{-4}-10^{-5}$) is quite similar.

\begin{figure}
    \centering
	\begin{subfigure}{1\linewidth}
    \includegraphics[trim={1cm 1cm 1cm 1cm},clip, width=0.95\linewidth]{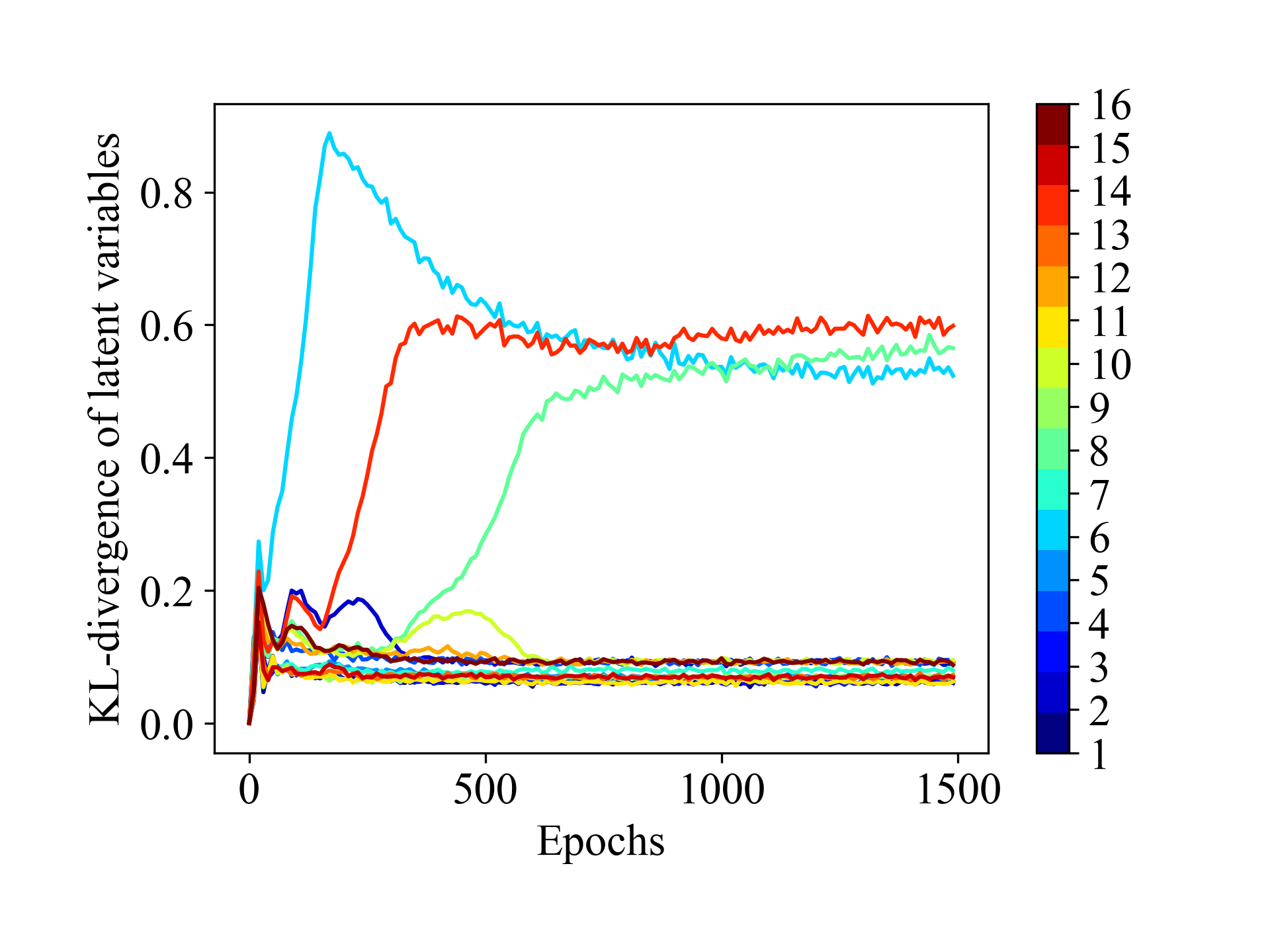}
    
	\caption{KL-divergence of latent variables in EVGAE}
	\label{fig:evgae-example-kld-16-8-1}
	\end{subfigure}
	\begin{subfigure}{1\linewidth}
    \includegraphics[trim={1cm 1cm 1cm 0.5cm},clip, width=0.95\linewidth]{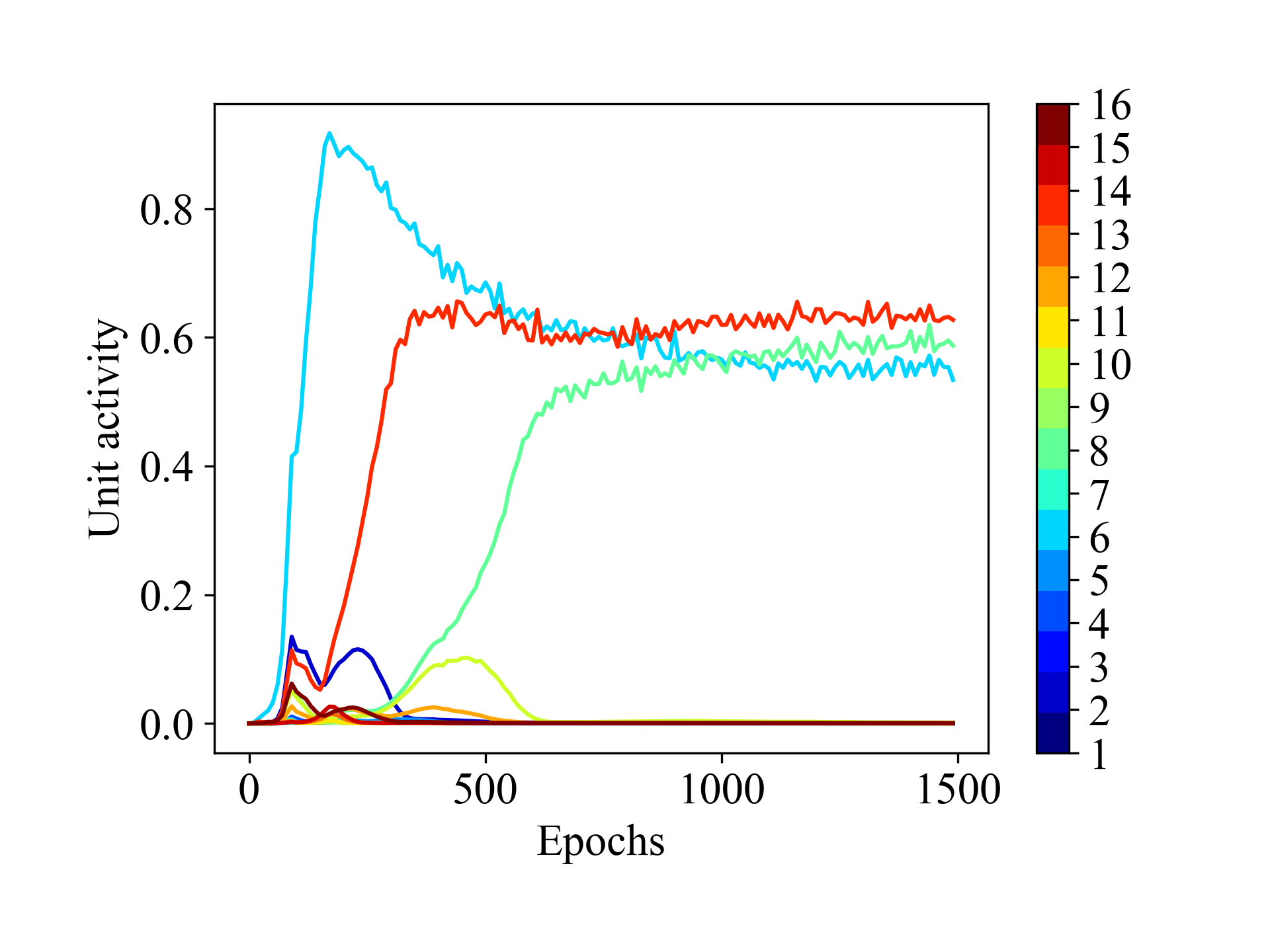}	\caption{Unit activity of 16 hidden units of EVGAE}
	\label{fig:evgae-example-activity}
	\end{subfigure}
	\caption{Three hidden units are active and KL-divergence of corresponding latent variables is quite low compared to Fig.~\ref{fig:vgae-beta-kld}, indicating a good matching of learnt distribution with prior, consequently improving the generative ability of the model.}
	\label{fig:evgae-example-16-8-1}
\end{figure}

\subsection{EVGAE: Over-pruning and Generative Ability}
We now show the learning behavior of EVGAE model on our running example of Cora dataset.
We select 8 epitomes, each dictating three hidden units to be active. The configuration is shown in Fig.~\ref{epitomes-example}. 
Fig.~\ref{fig:evgae-example-16-8-1} shows the evolution of KL-divergence and unit activity during training of EVGAE model. 
By comparing this figure with pure VGAE (Fig.~\ref{fig:vgae-normal}), we can observe that EVGAE has more active hidden units. This demonstrates that our model is better than pure VGAE at mitigating the over-pruning issue.

On the other hand, if we compare it to VGAE\cite{vgae}(Fig.~\ref{fig:vgae-beta}), we observe EVGAE have less active units in comparison. 
But KL-divergence of the latent variables for VGAE is greater than 1 for all the latent variables (Fig.~\ref{fig:vgae-beta-kld}). 
This implies that the latent distribution is quite different from the prior distribution (standard gaussian).
In contrast, we observe from Fig.~\ref{fig:evgae-example-kld-16-8-1} that EVGAE has KL-divergence around 0.1 for 13 latent variables and approximately 0.6 for remaining 3 latent variables. This reinforces our claim that VGAE achieves more active hidden units by excessively penalizing the KL-term responsible for generative ability.

In short, although EVGAE has less active units, the distribution matching is better compared to VGAE. VGAE is akin to GAE due to such low weightage to KL-term, i.e. $\beta$ = 0.0003. 

\begin{figure}[t]
	\vspace{0.2cm}
	\centering
	\begin{subfigure}{0.97\linewidth}
% 		\centering
		\begin{tikzpicture}
			\begin{axis}[
					width=0.9\linewidth,
					height=6.5cm,
					legend pos=north west,
					xlabel=Dimensions,
					xtick=data,
					ylabel=Active units,
					xticklabels from table={\channelskl}{Channels}
				] 
				\addplot[green,thick,mark=square*] table [y=vgaeau,x=X]{\channelskl};
				\addlegendentry{Pure VGAE ($\beta$ = 1)}
				\addplot[orange,thick,mark=square*] table [y=bvgaeau,x=X]{\channelskl};
				\addlegendentry{VGAE ($\beta$ = 0.0003 \cite{vgae})}
				\addplot[red,thick,mark=square*] table [y=evgaeau,x=X]{\channelskl};
				\addlegendentry{EVGAE}
			\end{axis}
		\end{tikzpicture}
		\caption{Active hidden units with varying latent space dimensions}%%Active hidden units with varying latent space dimensions}
		\label{fig:channels-au}
	\end{subfigure}
	\begin{subfigure}{0.97\linewidth}
% 		\centering
		\begin{tikzpicture}
			\begin{axis}[
					width=0.9\linewidth,
					height=6.5cm,
					xlabel=Dimensions,
					xtick=data,
					ylabel=Average KL-divergence per active unit,
					xticklabels from table={\channelskl}{Channels},
					yticklabel style={
						/pgf/number format/fixed,
						/pgf/number format/precision=5
					},
					scaled y ticks=false
				]						
				\addplot[green,thick,mark=square*] table [y=bvgaekl,x=X]{\channelskl};
				\addlegendentry{Pure VGAE ($\beta$ = 1)}
				\addplot[orange,thick,mark=square*] table [y=vgaekl,x=X]{\channelskl};
				\addlegendentry{VGAE ($\beta$ = 0.0003 \cite{vgae})}
				\addplot[red,thick,mark=square*] table [y=evgaekl,x=X]{\channelskl};
				\addlegendentry{EVGAE}
			\end{axis}
		\end{tikzpicture}
		\caption{}%{Average KL-divergence per active unit for different latent space dimensions}
		\label{fig:channels-kl}
	\end{subfigure}
	\caption{Effect of changing latent space dimensions on active units and their KL-divergence. It can be observed that EVGAE has more active units compared to VGAE, and with better generative ability}
	\label{fig:latent}
\end{figure}
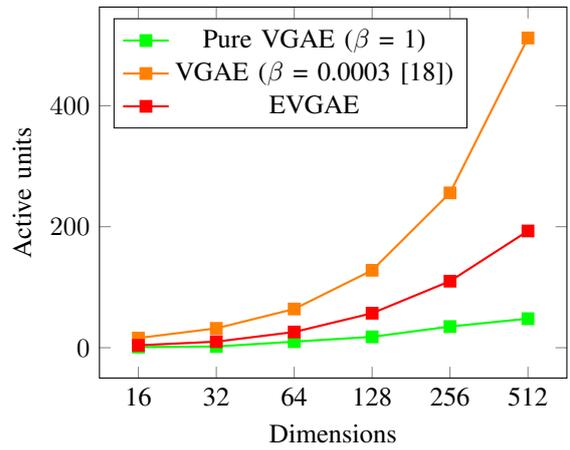
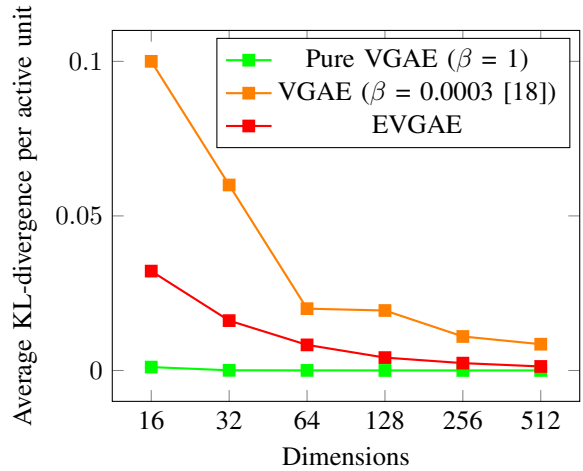

\subsection{Impact of Latent Space Dimension}%{Ablation Study}
We now look at the impact of latent space dimension on the number of active units and average KL-divergence per active unit.
We plot the active units for dimensions $D \in \{16, 32, 64, 128, 256, 512\}$. 
Fig.~\ref{fig:latent} presents an overview of this impact on our running example (Cora dataset). 
For all values of $D$, the number of epitomes is set to $\frac{D}{2}$ and one unit is allowed to overlap with neighboring epitomes. Similar to the configuration in Fig.~\ref{epitomes-example} for $D=16$. 
It is to be noted that we kept the same configuration of epitomes for consistency reasons.
Choosing a different configuration of epitomes does not affect the learning behavior of EVGAE. \ua{Due to space constraints, we do not include this analysis.}

It can be observed that the number of active units is quite less compared to the available units for VGAE with $\beta = 1$ (pure VGAE). Concretely, for $D=512$ only 48 units are active.
This shows that the over-pruning problem persists even in high dimensional latent space.

Now we observe the behavior of VGAE with $\beta = N^{-1}$ as proposed by Kipf and Welling\cite{vgae}, where $N$ denotes the number of nodes in the graph.
All the units are active irrespective of the dimension of latent space. 
In the case of EVGAE, the number of active units is in between the two. i.e. we are able to mitigate the over-pruning without sacrificing the generative ability ($\beta = 1$). 
This results in better performance in graph analysis tasks as shown in table \ref{tab:link-prediction-res}. 

To demonstrate that EVGAE achieves better distribution matching than VGAE, we compare the average KL-divergence of active units for different latent space dimensions. 
Only active units are considered when averaging the KL-divergence because the inactive units introduce a bias towards zero in the results. 
Fig.~\ref{fig:channels-kl} shows how the distribution matching varies as we increase the number of dimensions. 
We note that when $\beta=1$, the average KL-divergence for active units is still quite small, indicating a good match between learned latent distribution and the prior. 
Conversely, when $\beta = N^{-1}$ the average KL-divergence per active unit is quite high. This supports our claim that original VGAE\cite{vgae} learns a latent distribution which is quite different from the prior. Thus, when we generate new samples from standard gaussian distribution and pass it through the decoder, we get quite different output than the graph data used for training. 
In the case of EVGAE, the KL divergence is quite closer to the prior compared to VGAE. For $D=512$, it is almost similar to the case with $\beta = 1$.
\section{Conclusion}
In this paper we looked at the issue of over-pruning in variational graph autoencoder. We demonstrated that the way VGAE \cite{vgae} deals with this issue results in a latent distribution which is quite different from the standard gaussian prior. 
We proposed an alternative model based approach EVGAE that mitigates the problem of over-pruning by encouraging more latent variables to actively play their role in the reconstruction.
EVGAE also has a better generative ability than VGAE\cite{vgae} i.e. better matching between learned and prior distribution.
Moreover, EVGAE performs comparable or slightly better than the popular methods for the link prediction task. 

\section*{Acknowledgment}
This work has been supported by 
% an anonymous fund.
the Bavarian Ministry of Economic Affairs, Regional Development and Energy through the \emph{WoWNet} project  IUK-1902-003// IUK625/002.

% \section*{References}
\bibliographystyle{plain}
\bibliography{main}

% \begin{thebibliography}
% \end{thebibliography}

\end{document}